\documentclass[10pt,twocolumn,letterpaper]{article}

\usepackage[pagenumbers]{cvpr} 

\usepackage[dvipsnames]{xcolor}

\usepackage{graphicx}
\usepackage{diagbox}
\usepackage[pagebackref=true,breaklinks=true,colorlinks,bookmarks=false]{hyperref}
\usepackage{amsmath}
\usepackage{amssymb}
\usepackage[labelformat=simple]{subcaption}
\usepackage{multirow}
\usepackage{multicol}
\usepackage{diagbox}

\newcommand{\R}{\mathbb{R}}
\renewcommand{\t}{\boldsymbol{t}}

\def\paperID{*****} 
\def\confName{CVPR}
\def\confYear{2024}

\title{S3Aug: Segmentation, Sampling, and Shift for Action Recognition}

\author{Taiki Sugiura, Toru Tamaki\\
Nagoya Institute of Technology\\
Nagoya, Japan
}

\begin{document}
\maketitle

\begin{abstract}
    Action recognition is a well-established area of research in computer vision. In this paper, we propose S3Aug, a video data augmenatation for action recognition. Unlike conventional video data augmentation methods that involve cutting and pasting regions from two videos, the proposed method generates new videos from a single training video through segmentation and label-to-image transformation. Furthermore, the proposed method modifies certain categories of label images by sampling to generate a variety of videos, and shifts intermediate features to enhance the temporal coherency between frames of the generate videos. Experimental results on the UCF101, HMDB51, and Mimetics datasets demonstrate the effectiveness of the proposed method, paricularlly for out-of-context videos of the Mimetics dataset.

\end{abstract}

\section{Introduction}

Action recognition is an active area of research in computer vision and is used in a variety of applications. A major difficulty in developing action recognition methods is the need for a large amount of training data. To address this, several large datasets have been proposed~\cite{Kuehne_ICCV2011_HMDB51,Soomro_arXiv2012_UCF101,kay_arXiv2017_kinetics400,Goyal_2017ICCV_ssv2}.

In certain tasks, it can be hard to collect a large number of videos. To address this issue, data augmentation has been employed~\cite{Cauli_FI2022_video_augmentation_survey}. This technique involves virtually increasing the number of training samples by applying geometric transformations, such as vertical and horizontal flip, or image processing, such as cropping a part of one image and pasting it onto another.

Various data augmentation techniques have been proposed for both images~\cite{Shorten_BigData2019_augmentation_survey} and video tasks~\cite{Cauli_FI2022_video_augmentation_survey}. These video data augmentation methods are based on cutmix~\cite{Yun_2019_ICCV_cutmix} and copy paste~\cite{Ghiasi_2021_CVPR_Copy_Paste}, which involve cutting (or copying) regions of two videos to create a new video.
However, these approaches have two drawbacks. First, the spatio-temporal continuity of the actor in the video may be compromised. Unlike general image recognition tasks, the region of the actor is essential for action recognition, and thus simple extensions of cutmix or copy-paste are not suitable since the actor region may be cut off or obscured by augmentation. Second, action recognition datasets are known to have considerable dataset biases~\cite{Jihoon_NEURIPS2022_Action_swap}, therefore, simple augmentation does not address the issue of out-of-context (or out-of-distribution) samples.

Therefore, in this paper, we propose an alternative to cutmix-based data augmentation methods, called \emph{S3Aug} (\underline{S}egmentation, category \underline{S}ampling, and feature \underline{S}hift for video \underline{Aug}mentation). This method produces multiple videos from a single training video while maintaining the semantics of the regions by using panoptic segmentation and image translation.
We evaluated the effectiveness of our proposed method using two well-known action recognition datasets, UCF101~\cite{Soomro_arXiv2012_UCF101} and HMDB51~\cite{Kuehne_ICCV2011_HMDB51}. Furthermore, we evaluated its performance on out-of-context samples with the Mimetics data set~\cite{Weinzaepfel_IJCV2021_Mimetics_dataset}.

\begin{figure*}[t]
    \centering
    \includegraphics[width=.8\linewidth]{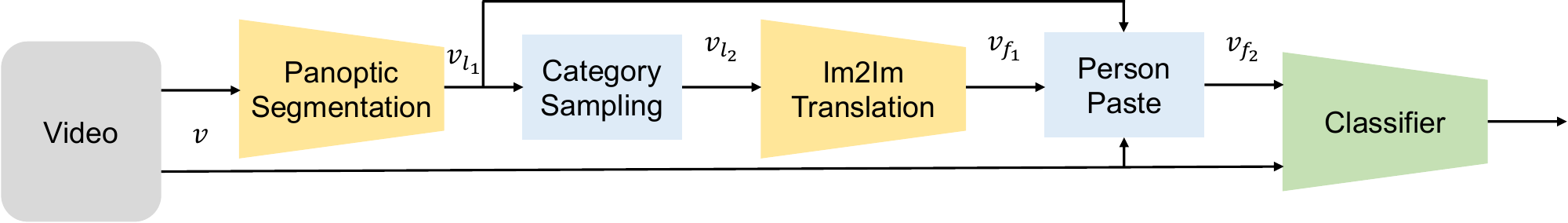}

    \caption{A schematic diagram of the proposed S3Aug. The green component is a classifier that is trained, while the yellow components are pre-trained segmentation and image translation components. The blue components are non-training processes.
    }
    \label{Fig:model overview}

\end{figure*}

\section{Related work}

Action recognition is a long-standing and significant area of study in computer vision
\cite{Hutchinson_IEEEAccess2021_Action_Recognition_Survey,Kong_IJCV2022_Action_Recognition_Survey,Ulhaq_arXiv2022_Transformers_Action_Recognition_Survey}, with a variety of models being proposed, including CNN-based~\cite{Feichtenhofer_CVPR2020,feichtenhofer_ICCV2019} and Transformer-based~\cite{Arnab_2021_ICCV_ViVit,Bertasius_ICML2021_TimeSformer}.

For this data-demanding task, video data augmentation has been proposed~\cite{Cauli_FI2022_video_augmentation_survey}. The main approach is cutmix~\cite{Yun_2019_ICCV_cutmix} and copy-paste~\cite{Ghiasi_2021_CVPR_Copy_Paste}, which cut (or coyp) a random rectangle or actor region from frames of one video and paste it onto frames of the other video. This approach is used by VideoMix~\cite{Yun_arXiv2020_VideoMix}, ActorCutMix~\cite{Zou_CVIU2022_ActorCutMix}, and ObjectMix~\cite{Kimata_MMAsia2022_ObjectMix}, however, it has the issue of the spatial and temporal discontinuity of the actor regions. To address this issue, Learn2Augment~\cite{Gowda_ECCV2022_Learn2Augment} and Action-Swap~\cite{Jihoon_NEURIPS2022_Action_swap} generate a background image by utilizing inpainting to remove the extracted actor regions from one video frame, and then paste the actors extracted from the other video frame onto the background image.

Another issue is background bias~\cite{Jihoon_NEURIPS2022_Action_swap,Weinzaepfel_IJCV2021_Mimetics_dataset,He_ECCVW2016_Action_Recognition_Without_Human}, where models tend to heavily rely on cues in appearances of the scene (e.g., background or object) and fail to predict the actions of out-of-context samples. To address this, some simple methods generate various videos from samples in the given dataset. Action Data Augmentation Framework~\cite{Wu_IJCNN_2019_GAN_aug_Action_Recognition} stacks the generated still images, which does not produce a video with appropriate variations. Self-Paced Selection~\cite{Zhang_ACMMM2020_GAN_aug_Action_Recognition} treats a video as a single ``dynamic image'', resulting in the loss of temporal information. Our approach is similar in spirit but instead uses segmentation as a guide to generate video frames to maintain the semantics of the original source video.

Note that generating videos is still a difficult task
despite advances in generative models
such as GAN~\cite{Jabbar_ACMSurvey2021_GAN_survey,Goodfellow_NIPS2014_GAN,Yi_MedIA2019_GAN_medical_imaging_survey}
and diffusion models~\cite{Rombach_2022_CVPR_stable_diffusion,Ramesh_ICML2021_DALL-E}.
There have been some attempts to generate videos using diffusion models~\cite{Ho_NEURIPS2022_Video_Diffusion_Models,Luo_2023_CVPR_VideoFusion} and
GPT~\cite{Yan_arXiv2011_VideoGPT},
but they require specific prompts to control the content of the videos, which is an ongoing exploration.
On the contrary, our approach produces video frames from segmented label frames, similar to Vid2Vid~\cite{Wang_NeurIPS2018_vid2vid} and the more recent ControlNet~\cite{Zhang_arXiv2023_ControlNet}.
However, these methods are computationally expensive and are not suitable for this study.
Therefore, we use a GAN-based method~\cite{Park_2019_CVPR_SPADE} as frame-wise image translation,
which is a compromise between speed and computational cost.
Frame-wise processes are known
to produce temporally incoherent results,
so we propose the shit feature, which was originally proposed
for lightweight action recognition models~\cite{Zhang_ACMMM2021_TokenShift,Lin_2019ICCV_TSM,Hashiguchi_2022_ACCVW_MSCA,Wang_arXiv2022_ShiftViT}.

\section{Method}

This section begins by providing an overview of the proposed S3Aug (Fig.\ref{Fig:model overview}), followed by a description of key components such as category sampling and feature shift.

An input video clip $v \in \R^{T \times 3 \times H \times W}$ is a sequence of $T$ frames $v(t) \in \R^{3 \times H \times W}, t=1,\ldots,T$, where $H, W$ are the height and width of the frame. $y \in \{0, 1\}^{L_a}$ is a one-hot vector of the action label, with $L_a$ being the number of action categories.

First, we apply panoptic segmentation to each frame $v(t)$ to obtain the corresponding label image $v_{l_1}(t) \in \{0, 1, \ldots, L_s\}^{H \times W}$,
where $L_s$ is the number of segmentation categories.
In addition, an instance map $I(t)\in \{0, 1, \ldots, N(t)\}^{H \times W}$ is obtained, which assigns a unique value to each detected instance in the frame,
with $N(t)$ being the number of instances detected in the frame.
To obtain another label image $v_{l_2}(t)$, some of the pixels in $v_{l_1}(t)$ are replaced by the proposed category sampling (which will be discussed in sec.\ref{sec:Category sampling}).

Then the label image $v_{l_2}(t)$ and the instance map $I(t)$
are used to generate the image $v_{f_1}(t) \in \R^{3 \times H \times W}$ using the image translation with feature shift (sec.\ref{sec:Feature shift}),
and then the actors' regions are pasted to generate the final frame $v_{f_2}(t)$.

\subsection{Category sampling}
\label{sec:Category sampling}

In the image translation stage, the frame generated from a given frame $v(t)$ would remain the same for each epoch unless the method has a stochastic mechanism. Diffusion models have this, but we opted for a deterministic GAN for this step for the aforementioned reason.
To introduce variability in the generated frames even with deterministic methods,
we propose replacing the segmentation label category in the label images with a different category. We call this process \emph{category sampling}.

This is similar to introducing noise into the latent variable~\cite{Zhu_NIPS2017_BicycleGAN} to create a variety of images; however, it is difficult to maintain frame-to-frame temporal consistency. On the other hand, simply replacing the categories of labeled images can be done very quickly, and it is possible to maintain temporal continuity between frames when the categories are replaced in the same way for all frames.

The importance of which categories are replaced is a key factor in this work.
We use the COCO dataset~\cite{Lin_ECCV2014_COCO}, which is the de fact standard for segmentation tasks.
The idea is to maintain objects in the scene that are essential for understanding the actions and people-object interactions.
Therefore, the COCO things~\cite{Lin_ECCV2014_COCO} categories, including the person class, are kept as is, while the COCO stuff~\cite{Caesar_2018_CVPR_COCO_stuff} categories are replaced.
In the following, we propose two methods,
random sampling (for random categories)
and semantic sampling (for semantically similar categories).

\begin{figure}[t]
    \centering
    \begin{minipage}[t]{0.22\linewidth}
        \includegraphics[width=\linewidth]{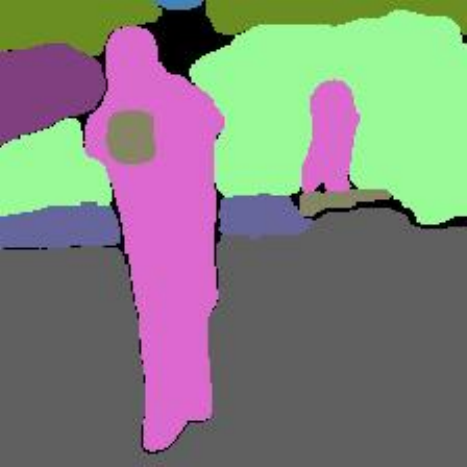}
        \subcaption{}
        \label{fig:v_l_1}
    \end{minipage}
    \hspace{2em}
    \begin{minipage}[t]{0.45\linewidth}
        \includegraphics[width=\linewidth]{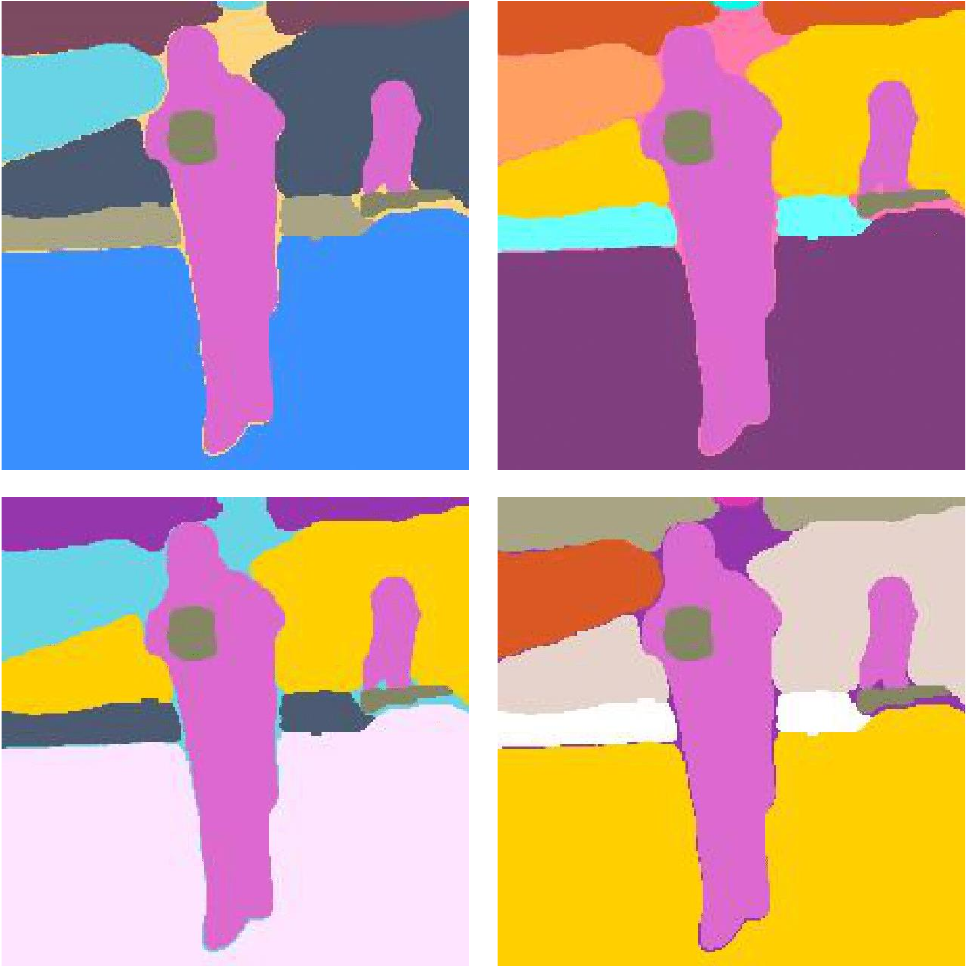}
        \subcaption{}
        \label{fig:v_l_2}
    \end{minipage}

    \caption{Category sampling.
        (a) A frame of a label video $v_{l_1}$,
        (b) corresponding several frames of $v_{l_2}$.
    }
    \label{fig:category sampling}
\end{figure}

\subsubsection{Random category sampling}

We use a segmentation model
pre-trained on the COCO panoptic segmentation~\cite{Kirillov_2019_CVPR_panoptic_segmentation}.
The category set of segmentation
$\{0, 1, \ldots, L_s = 200\}$ is consists of
the unlabeled class $\{0\}$,
the things class set $L_\mathrm{things} = \{1, \ldots, 91\}$,
the stuff class set $L_\mathrm{stuff} = \{ 92, \ldots, 182\}$,
and
the merged stuff class set $L_\mathrm{mstuff} = \{183, \ldots, 200\}$%
\footnote{
    \url{https://cocodataset.org/\#panoptic-eval}
}.

For each video, we use a permutation $\sigma$ that represents replacement sampling, randomly replacing categories of the COCO stuff and merged stuff to one of the categories of the COCO stuff
\begin{align}
    \sigma =
    \begin{pmatrix}
        92         & \cdots & 200         \\
        \sigma(92) & \cdots & \sigma(200)
    \end{pmatrix},
\end{align}
and each $\sigma(c)$ is sampled by
\begin{align}
    \sigma(c)\sim \mathrm{Unif}(L_\mathrm{stuff}) \quad \forall c \in L_\mathrm{stuff} \cup L_\mathrm{mstuff},
\end{align}
where $\mathrm{Unif}$ is a uniform distribution.
Note that
$\{\sigma(c) \ | \ c \in L_\mathrm{stuff} \cup L_\mathrm{mstuff} \}
    \subseteq L_\mathrm{stuff}$
holds due to the replacement.
The same permutation is used for each video, and it is applied to all pixels in all frames to create a new label image;
\begin{align}
    v_{l_2}(x, y, t) & = \sigma (v_{l_1}(x, y, t)),
\end{align}
where $v(x,y,t)$ denotes the pixel values of the corresponding frame of the video $v$.

\subsubsection{Semantic category sampling}

Rather than randomly selecting categories, a category sampling that takes into account the similarity between them by using word embedding is expected to generate more realistic frames than simply substituting categories randomly. We call this sampling semantic category sampling.

First, the category name $w_c$ of each stuff category $c \in L_\mathrm{stuff}$ is encoded into an embedding $\t_c$.
Then, we compute the cos similarity of the embedding $\t_c$ of category $c$ to the embedding $\t_{c'}$ of other categories $c'$,
\begin{align}
    p(c'|c) & =
    \frac{\exp(\t_c^T \t_{c'})}{\sum_{i \in L_\mathrm{stuff}} \exp(\t_c^T \t_i)},
\end{align}
and sample a new category
\begin{align}
    c' \sim p(c'|c) \quad \forall c \in L_\mathrm{stuff}.
\end{align}
Similarly to random category sampling, we fix $\sigma(c) = c'$ for all frames of each video
to obtain a new label image.

\subsection{Feature shift}
\label{sec:Feature shift}

It is known that frame-wise processing often results in
temporal incoherency; the resulting video exhibit artifacts such as flickering between frames.
In this study, we propose the use of feature shift,
which has been proposed to give the ability of temporal modeling to frame-wise image recognition models~\cite{Lin_2019ICCV_TSM,Zhang_ACMMM2021_TokenShift,Hashiguchi_2022_ACCVW_MSCA,Wang_arXiv2022_ShiftViT}.
This approach inserts feature shift modules inside or between layers of a 2D CNN or transformer model to swap parts of features between consecutive frames.
We use feature shift to enhance coherency between frames.

A typical structure of image translation models consists of a combination of an encoder and a decoder, both of which are composed of multiple blocks.
Assuming that there are no skip connections across blocks,
we write the $\ell$-th decoder block as follows
\begin{align}
    z_{\ell} & = D^\ell (z_{\ell - 1}),
\end{align}
where $z_{\ell} \in R^{T \times C_\ell \times H_\ell \times W_\ell}$ are intermediate features and $C_\ell, H_\ell, W_\ell$ are the number of channels, height and width.
In this work, we insert a feature shift module between the decoder blocks as follows;
\begin{align}
    z'_{\ell} & = D^\ell (z_{\ell - 1})       \\
    z_{\ell}  & = \mathrm{shift} (z'_{\ell}).
\end{align}
Let $z_{\ell}(t, c) \in \R^{H_\ell \times W_\ell}$
be the channel $c$ of $t$-th frame of $z_{\ell}$,
then the shift module
can be represented as follows~\cite{Hashiguchi_2022_ACCVW_MSCA};
\begin{align}
    z_{\ell}(t, c)
     & =
    \begin{cases}
        z'_{\ell}(t-1, c), & \text{$1 < t \le T, 1 \le c < C_b$}       \\
        z'_{\ell}(t+1, c), & \text{$1 \le t < T, C_b \le c < C_b+C_f$} \\
        z'_{\ell}(t,   c), & \text{$\forall t, C_b+C_f \le c \le C$}   \\
    \end{cases}.
\end{align}
This means that the first $C_b$ channels at time $t$
are shifted backward to time $t-1$,
and
the next $C_f$ channels are shifted forward to time $t+1$.

\begin{figure}[t]
    \centering

    \begin{minipage}[t]{\linewidth}
        \includegraphics[width=\linewidth]{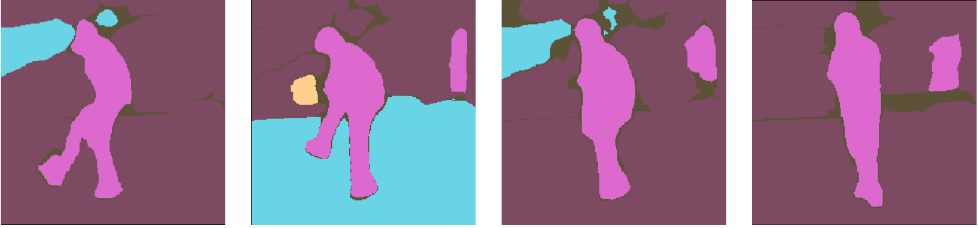}
        \subcaption{}
    \end{minipage}

    \vspace{0.2cm}

    \begin{minipage}[t]{\linewidth}
        \includegraphics[width=\linewidth]{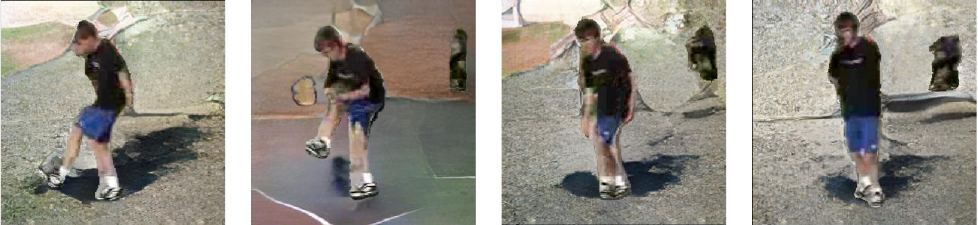}
        \subcaption{w/o shift}
    \end{minipage}

    \vspace{0.2cm}

    \begin{minipage}[t]{\linewidth}
        \includegraphics[width=\linewidth]{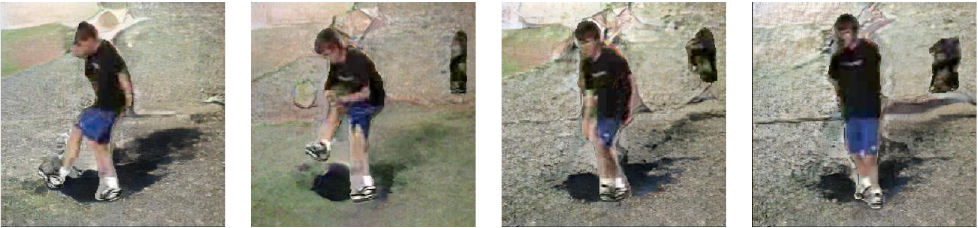}
        \subcaption{w shift}
    \end{minipage}

    \caption{Effect of feature shift.
        (a) Label images $V_{l_2}$, and
        (b) corresponding generated frames without feature shift and
        (c) with feature shift.
    }
    \label{fig:feature shift}
\end{figure}

Note that we used a pre-trained image translation model%
~\cite{Park_2019_CVPR_SPADE} in which shifting was not considered.
However, it is expected to contribute to the reduction of artifacts between frames,
as shown in Figure \ref{fig:feature shift}.

\subsection{Person paste}

\begin{figure}[t]
    \centering
    \begin{minipage}[t]{0.2\linewidth}
        \includegraphics[width=\linewidth]{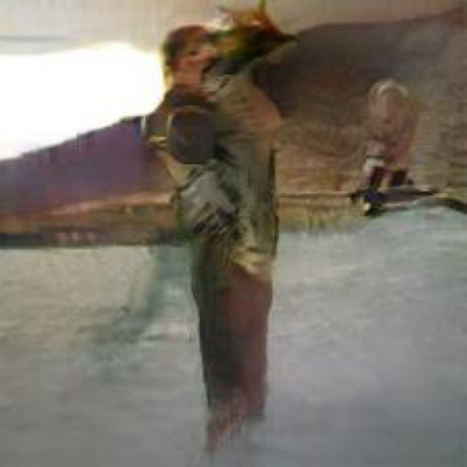}
        \subcaption{}
    \end{minipage}
    \hspace{2em}
    \begin{minipage}[t]{0.2\linewidth}
        \includegraphics[width=\linewidth]{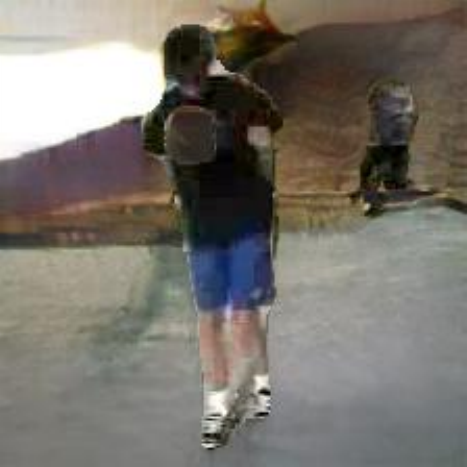}
        \subcaption{}
    \end{minipage}

    \caption{Example of person paste.
        (a) From the labeled moving image $v_{l_2}$ of Figure \ref{fig:v_l_2}
        The generated video $v_{f_1}$ and
        (b) Video image with person area pasted on it $V_{F_2}$.
    }
    \label{fig:person paste}
\end{figure}

A pre-trained image translation model might work in general; however, it does not guarantee to generate plausible actors that are import to action recognition.
Therefore, we use the actor regions in the original video frame and paste them into the generated frame as shown in Fig. \ref{fig:person paste}.
\begin{align}
    v_{f_2}(x,y,t) & =
    \begin{cases}
        v(x,y,t)       & v_{l_1}(x,y,t) = \text{``person''} \\
        v_{f_1}(x,y,t) & \text{otherwise}                   \\
    \end{cases}
\end{align}

\section{Experimental results}

We evaluate the proposed S3Aug
with two commonly used action recognition datasets
and an out-of-context dataset.
We also compare it with the conventional methods.

\subsection{Settings}

\subsubsection{Datasets}

UCF101~\cite{Soomro_arXiv2012_UCF101} is a motion recognition dataset of 101 categories of human actions, consisting of a training set of approximately 9500 videos and a validation set of approximately 3500 videos.

HMDB51~\cite{Kuehne_ICCV2011_HMDB51} consists of a training set of about 3600 videos and a validation set of about 1500 videos.
HMDB51 is a motion recognition dataset of 51 categories of human motions.

Mimetics~\cite{Weinzaepfel_IJCV2021_Mimetics_dataset} is an evaluation-only dataset consisting of 713 videos with 50 categories, which is a subset of the category of Kinetics400~\cite{kay_arXiv2017_kinetics400}. Videos are out-of-context that does not align the usual context of action recognition, such as surfing in a room or bowling on a football pitch. After training on 50 categories of the Kinetics400 training set, of which videos are normal context, we evaluated on the 50 categories of the Mimetics dataset.

\subsubsection{Model}

Mask2Former~\cite{Cheng_CVPR2022_mask2former} pre-trained on the COCO Panoptic segmentation~\cite{Kirillov_2019_CVPR_panoptic_segmentation}
(80 things, 36 stuff, 17 other, 1 unlabeled classes)
was used for the segmentation of each frame of the video.

For the image translation model, we used SPADE~\cite{Park_2019_CVPR_SPADE} pre-trained on the COCO stuff%
~\cite{Caesar_2018_CVPR_COCO_stuff}
(80 things, 91 stuff, 1 unlabelled classes).
The number of channels $C_b, C_f$ to be shifted was set to $C_\ell/8$ in every layer, where $C_\ell$ is the number of channels of the feature $z_\ell$.

A pretrained BERT~\cite{Devlin-ACL2019-BERT} was used for the word embedding model.
For action recognition, we used X3D-M~\cite{Feichtenhofer_2020CVPR_X3D} pre-trained on Kinetics400~\cite{kay_arXiv2017_kinetics400}.

In the experiment, only the action recognition model, X3D-M, was fine-trained, while the other models were pre-trained and fixed.
For feature shift, a shift module was inserted into each decoder block of the pre-trained SPADE with the weights fixed.

\subsubsection{Training and evaluation}

We followed a standard training setting. We randomly sampled 16 frames from a video to form a clip, randomly determined the short side of the frame in the $[256, 320]$ pixel range, resized it to preserve the aspect ratio, and randomly cropped a $224 \times 224$ patch.
Unless otherwise noted, the number of training epochs was set at 10, batch size at 2, and learning rate at 1e-4 with Adam optimizer~\cite{Kingma_ICLR2015_Adam}.

In validation, we used the multiview test~\cite{Wang_2018CVPR_NonLocal_multiview}
with 30 views;
three different crops from 10 clips randomly sampled.

We applied the proposed method to each batch with
probability $0 \le p \le 1$.
In the experiment, the performance was evaluated from $p=0$ to $p=1$ in increments of 0.2.
Note that $p=0$ is equivalent to no augmentation.

\subsection{Effects of components}

Table \ref{tab:comparison} shows
the effect of category sampling, feature shift (fs),
and person paste (pp).
Note that the results are identical for $p=0.0$.

The first row shows results without any proposed modules.
Performance decreases when $p > 0$, demonstrating that
a simple image translation only does not work as a video data augmentation.

The second row shows the result of the person paste,
showing that the person paste consistently improves performance for all $p$ values.
The performance decrease for large $p$ is less significant than when the person paste is used,
indicating that the effect of the person paste is more pronounced.

Without feature shift,
random sampling looks slightly better than semantic sampling
as shown in the third and fourth rows.
However, as shown in the last two rows,
semantic category sampling shows better than or comparable performance
with feature shift.
The best performances of the random and semantic category sampling are the same
at $p=0.2$, while the semantic category sampling performs slightly better
for other values of $p$.

Note that in all settings, performance decreases as $p$ increases and, in particular,
performance decreases significantly for $p \ge 0.6$, regardless of which setting was used.
This indicates that the augmented samples clearly change the content of the frames
and that too much augmentation does not help the model to be generalized.

\begin{table}[t]
    \begin{center}
        \caption{Evaluation of top-1 performance on the UCF101 validation set
            for random (r) and semantic (s) category sampling (cs), feature shift (fs), and person paste (pp).
        }
        \label{tab:comparison}

        \begin{tabular}{r@{\, }c@{\, }c@{\, }|c@{\, }c@{\, }c@{\, }c@{\, }c@{\, }c}
            cs & fs         & pp         & 0.0   & 0.2            & 0.4   & 0.6   & 0.8   & 1.0   \\ \hline
               &            &            & 93.68 & 92.71          & 93.38 & 90.88 & 89.20 & 74.41 \\
               &            & \checkmark & 93.68 & 94.15          & 94.10 & 91.66 & 90.60 & 85.28 \\
            r  &            & \checkmark & 93.68 & 94.04          & 92.99 & 92.93 & 91.85 & 82.93 \\
            s  &            & \checkmark & 93.68 & 93.99          & 93.63 & 92.96 & 89.83 & 81.71 \\
            r  & \checkmark & \checkmark & 93.68 & \textbf{94.26} & 93.85 & 92.93 & 91.86 & 82.93 \\
            s  & \checkmark & \checkmark & 93.68 & \textbf{94.26} & 93.54 & 92.77 & 90.49 & 83.73 \\
        \end{tabular}
    \end{center}
\end{table}

\subsection{Comparisons}

The comparison with VideoMix~\cite{Yun_arXiv2020_VideoMix} and ObjectMix~\cite{Kimata_MMAsia2022_ObjectMix} on UCF101 and HMDB51 is shown in Table \ref{tab:comparison UCF HMDB}. The batch size was 16, which is the same as in the previous work~\cite{Kimata_MMAsia2022_ObjectMix}.

The results of the experiments vary depending on the randomness of the training and the augmentation applied.
Therefore, we ran each setting three times for each method, and the results are presented in a single cell, along with the average performance of the cell.

The proposed S3Aug performs competitively on UCF101 and significantly better on HMDB51, with an average of 78.88\% ($p=0.2$), which is 2 points higher than the best of VideoMix and ObjectMix. It is likely that a similar performance of the three methods is obtained on UCF101, as the data set is relatively easy to predict, and the state-of-the-art methods exceed 98\%~\cite{Wang_2023_CVPR_VideoMAE2}.

Generally, VideoMix and ObjectMix appear to be more effective when $p$ is larger (around 0.6), while S3Aug is most successful when $p$ is around 0.2. This discrepancy is due to the fact that the techniques generate videos in the same or different contexts.
VideoMix and ObjectMix generate new videos by utilizing two training videos, which share a similar context in terms of the background. On the other hand, S3Aug produces a video with a completely different background from the original video. We compare these methods in this paper, but our method is complementary to them, and thus a synergistic effect can be expected when they are used together.

\begin{table}[t]

    \centering

    \caption{Performance comparison of the proposed S3Aug with two previous work; VideoMix and ObjectMix.
        The top one is on the validation set of UCF101 and the bottom is HMDB51.
    }
    \label{tab:comparison UCF HMDB}

    {\small
        \begin{tabular}{r@{\, }|c@{\, }c@{\, }c@{\, }c@{\, }c@{\, }c}
            method    & 0.0   & 0.2   & 0.4   & 0.6   & 0.8   & 1.0   \\ \hline
                      & 93.40 & 93.18 & 93.49 & 93.60 & 92.96 & 92.66 \\
            VideoMix  & 93.68 & 93.51 & 93.82 & 93.65 & 93.99 & 92.85 \\
                      & 94.06 & 94.51 & 94.15 & 93.96 & 94.20 & 93.23 \\ \cline{2-7}
            avg       & 93.71 & 93.73 & 93.82 & 93.74 & 93.72 & 92.91 \\
            \hline
                      & 93.40 & 94.07 & 94.10 & 93.68 & 94.10 & 92.74 \\
            ObjectMix & 93.68 & 94.10 & 94.15 & 93.71 & 94.34 & 92.82 \\
                      & 94.06 & 94.20 & 94.37 & 94.76 & 94.48 & 93.76 \\ \cline{2-7}
            avg       & 93.71 & 94.12 & 94.21 & 94.05 & 94.31 & 93.11 \\
            \hline
                      & 93.40 & 93.29 & 93.21 & 92.07 & 90.19 & 83.54 \\
            S3Aug     & 93.68 & 94.04 & 93.54 & 92.77 & 90.49 & 83.73 \\
                      & 94.06 & 94.15 & 94.76 & 93.38 & 90.74 & 84.45 \\ \cline{2-7}
            avg       & 93.71 & 93.83 & 93.84 & 92.74 & 90.47 & 83.91 \\
            \hline
        \end{tabular}
    }

    \vspace{1em}

    {\small
        \begin{tabular}{r@{\, }|c@{\, }c@{\, }c@{\, }c@{\, }c@{\, }c}
            method    & 0.0   & 0.2   & 0.4   & 0.6   & 0.8   & 1.0   \\ \hline
                      & 74.22 & 75.11 & 74.33 & 76.67 & 74.56 & 74.42 \\
            VideoMix  & 76.83 & 76.50 & 76.89 & 76.89 & 75.56 & 75.50 \\
                      & 78.22 & 77.83 & 78.39 & 77.67 & 76.39 & 75.89 \\ \cline{2-7}
            avg       & 76.42 & 76.48 & 76.54 & 77.08 & 75.50 & 75.27 \\
            \hline
                      & 74.22 & 76.33 & 75.56 & 75.33 & 72.17 & 74.72 \\
            ObjectMix & 76.83 & 77.00 & 76.67 & 75.83 & 75.83 & 73.83 \\
                      & 78.22 & 77.33 & 78.39 & 77.67 & 76.39 & 75.89 \\ \cline{2-7}
            avg       & 76.42 & 76.89 & 76.87 & 76.28 & 74.80 & 74.81 \\
            \hline
                      & 74.22 & 77.00 & 77.72 & 76.17 & 72.22 & 70.11 \\
            S3Aug     & 76.83 & 79.81 & 77.89 & 76.50 & 75.83 & 73.94 \\
                      & 78.22 & 79.83 & 78.22 & 79.17 & 77.22 & 74.11 \\ \cline{2-7}
            avg       & 76.42 & 78.88 & 77.94 & 77.28 & 75.09 & 72.72 \\
            \hline
        \end{tabular}
    }

\end{table}

\subsection{Performance on out-of-context videos}

One of the motivations of the proposed S3Aug is to address the issue of
the background bias by generating various background while keeping the semantic layout of the action scene.
Table \ref{tab:comparison kinetics mimetics} shows the performance comparisons
of the proposed method and other two prior work.
The top table shows performances on the same 50 categories of Kinetics validation set,
which is in-context samples. Three methods are almost comparable while S3Aug is inferior
due to the reason mentioned above.

The motivation behind S3Aug is to tackle the problem of background bias by creating a variety of backgrounds while preserving the semantic layout of the action scene. Table \ref{tab:comparison kinetics mimetics} compares the performance of the proposed method with two prior works. The top table displays the results on the same 50 categories of the Kinetics validation set, which are in-context samples. All three methods are quite similar, however S3Aug is slightly weaker due to the previously mentioned reason.

\begin{figure*}[t]
    \centering

    \includegraphics[width=0.6\linewidth]{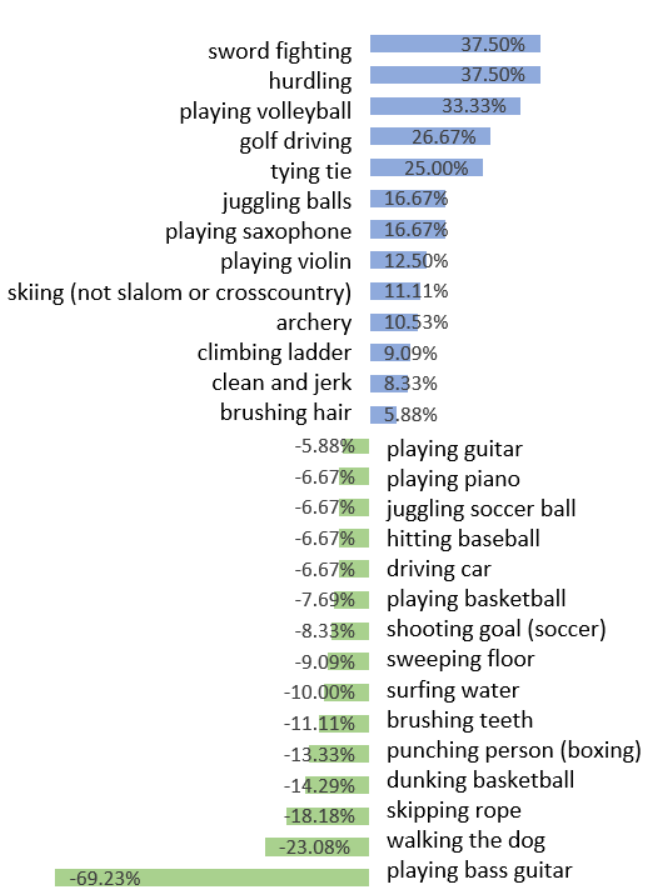}

    \caption{The score differences of 50 action categories of the Mimetics datasets when S3Aug is used and when it is not. Categories with no differences are not included in the comparison.
    }
    \label{fig:score differences}
\end{figure*}

The bottom table shows the results for the Mimetics dataset, which clearly demonstrate the superiority of the proposed method. This is likely due to the various background generated by the proposed method.
Figure \ref{fig:score differences} shows how socres of each category were improved or deteriorated when S3Aug is used
relative to the case when it is not used ($p=0.0$).
The top four categories are of sports, and training a model with generated various background may help to handle out-of-context videos of the Mimetics dataset.
The worst categories look involving objects (e.g. guitar, leash for dogs, rope) that are not included as a category of the COCO dataset, or are too small to be detected by the segmentation model.
This is a limitation of the proposed approach and using more sophisticated segmenation models or datasets with fine categories would be helpful.

\begin{table}[t]

    \centering

    \caption{Evaluation of top-1 performance on 50 categories of the Kinetics (top) and Mimetics (bottom) validation sets.}
    \label{tab:comparison kinetics mimetics}

    {\small
        \begin{tabular}{r@{\, }|c@{\, }c@{\, }c@{\, }c@{\, }c@{\, }c}
            method    & 0.0   & 0.2   & 0.4   & 0.6   & 0.8   & 1.0   \\ \hline
            VideoMix  & 81.99 & 79.88 & 78.75 & 79.60 & 79.72 & 78.38 \\
            ObjectMix & 81.99 & 78.30 & 78.55 & 78.30 & 77.41 & 79.07 \\
            S3Aug     & 81.99 & 81.63 & 80.54 & 79.60 & 77.22 & 66.72 \\
        \end{tabular}
    }

    \vspace{1em}

    {\small
        \begin{tabular}{r@{\, }|c@{\, }c@{\, }c@{\, }c@{\, }c@{\, }c}
            method    & 0.0   & 0.2   & 0.4   & 0.6            & 0.8   & 1.0   \\ \hline
            VideoMix  & 16.72 & 16.09 & 16.09 & 15.61          & 16.72 & 17.98 \\
            ObjectMix & 16.72 & 15.68 & 15.77 & 16.24          & 13.88 & 17.35 \\
            S3Aug     & 16.72 & 19.30 & 22.37 & \textbf{22.40} & 19.08 & 16.45 \\
        \end{tabular}
    }

\end{table}

\section{Conclusion}

In this study, we proposed S3Aug, a video data augmenatation for action recognition using segmentation, category sampling, image generation, and feature shift. The proposed method is different from conventional data augmentation methods that cut and paste object regions from two videos in that it generates a label video from a single video by segmentation and creates a new video by image translation. Experiments using UCF101 and HMDB51 have confirmed that UCF101 is effective as a data augmentation method to suppress overfit during training.

\section*{\uppercase{Acknowledgements}}

This work was supported in part by JSPS KAKENHI Grant Number JP22K12090.

    {
        \small
        \bibliographystyle{ieeenat_fullname}
        \bibliography{mybib}

\begin{thebibliography}{49}
\providecommand{\natexlab}[1]{#1}
\providecommand{\url}[1]{\texttt{#1}}
\expandafter\ifx\csname urlstyle\endcsname\relax
  \providecommand{\doi}[1]{doi: #1}\else
  \providecommand{\doi}{doi: \begingroup \urlstyle{rm}\Url}\fi

\bibitem[Arnab et~al.(2021)Arnab, Dehghani, Heigold, Sun, Lu\v{c}i\'c, and Schmid]{Arnab_2021_ICCV_ViVit}
Anurag Arnab, Mostafa Dehghani, Georg Heigold, Chen Sun, Mario Lu\v{c}i\'c, and Cordelia Schmid.
\newblock Vivit: A video vision transformer.
\newblock In \emph{Proceedings of the IEEE/CVF International Conference on Computer Vision (ICCV)}, pages 6836--6846, 2021.

\bibitem[Bertasius et~al.(2021)Bertasius, Wang, and Torresani]{Bertasius_ICML2021_TimeSformer}
Gedas Bertasius, Heng Wang, and Lorenzo Torresani.
\newblock Is space-time attention all you need for video understanding?
\newblock In \emph{Proceedings of the 38th International Conference on Machine Learning}, pages 813--824. PMLR, 2021.

\bibitem[Caesar et~al.(2018)Caesar, Uijlings, and Ferrari]{Caesar_2018_CVPR_COCO_stuff}
Holger Caesar, Jasper Uijlings, and Vittorio Ferrari.
\newblock Coco-stuff: Thing and stuff classes in context.
\newblock In \emph{Proceedings of the IEEE Conference on Computer Vision and Pattern Recognition (CVPR)}, 2018.

\bibitem[Cauli and Reforgiato~Recupero(2022)]{Cauli_FI2022_video_augmentation_survey}
Nino Cauli and Diego Reforgiato~Recupero.
\newblock Survey on videos data augmentation for deep learning models.
\newblock \emph{Future Internet}, 14\penalty0 (3), 2022.

\bibitem[Cheng et~al.(2022)Cheng, Misra, Schwing, Kirillov, and Girdhar]{Cheng_CVPR2022_mask2former}
Bowen Cheng, Ishan Misra, Alexander~G. Schwing, Alexander Kirillov, and Rohit Girdhar.
\newblock Masked-attention mask transformer for universal image segmentation.
\newblock 2022.

\bibitem[Chung et~al.(2022)Chung, Wu, and Russakovsky]{Jihoon_NEURIPS2022_Action_swap}
Jihoon Chung, Yu Wu, and Olga Russakovsky.
\newblock Enabling detailed action recognition evaluation through video dataset augmentation.
\newblock In \emph{Advances in Neural Information Processing Systems}, pages 39020--39033. Curran Associates, Inc., 2022.

\bibitem[Devlin et~al.(2019)Devlin, Chang, Lee, and Toutanova]{Devlin-ACL2019-BERT}
Jacob Devlin, Ming-Wei Chang, Kenton Lee, and Kristina Toutanova.
\newblock {BERT}: Pre-training of deep bidirectional transformers for language understanding.
\newblock In \emph{Proceedings of the 2019 Conference of the North {A}merican Chapter of the Association for Computational Linguistics: Human Language Technologies, Volume 1 (Long and Short Papers)}, pages 4171--4186, Minneapolis, Minnesota, 2019. Association for Computational Linguistics.

\bibitem[Feichtenhofer(2020{\natexlab{a}})]{Feichtenhofer_2020CVPR_X3D}
Christoph Feichtenhofer.
\newblock X3d: Expanding architectures for efficient video recognition.
\newblock In \emph{Proceedings of the IEEE/CVF Conference on Computer Vision and Pattern Recognition (CVPR)}, 2020{\natexlab{a}}.

\bibitem[Feichtenhofer(2020{\natexlab{b}})]{Feichtenhofer_CVPR2020}
Christoph Feichtenhofer.
\newblock X3d: Expanding architectures for efficient video recognition.
\newblock In \emph{Proceedings of the IEEE/CVF Conference on Computer Vision and Pattern Recognition (CVPR)}, 2020{\natexlab{b}}.

\bibitem[Feichtenhofer et~al.(2019)Feichtenhofer, Fan, Malik, and He]{feichtenhofer_ICCV2019}
Christoph Feichtenhofer, Haoqi Fan, Jitendra Malik, and Kaiming He.
\newblock Slowfast networks for video recognition.
\newblock In \emph{Proceedings of the IEEE/CVF international conference on computer vision}, pages 6202--6211, 2019.

\bibitem[Ghiasi et~al.(2021)Ghiasi, Cui, Srinivas, Qian, Lin, Cubuk, Le, and Zoph]{Ghiasi_2021_CVPR_Copy_Paste}
Golnaz Ghiasi, Yin Cui, Aravind Srinivas, Rui Qian, Tsung-Yi Lin, Ekin~D. Cubuk, Quoc~V. Le, and Barret Zoph.
\newblock Simple copy-paste is a strong data augmentation method for instance segmentation.
\newblock In \emph{Proceedings of the IEEE/CVF Conference on Computer Vision and Pattern Recognition (CVPR)}, pages 2918--2928, 2021.

\bibitem[Goodfellow et~al.(2014)Goodfellow, Pouget-Abadie, Mirza, Xu, Warde-Farley, Ozair, Courville, and Bengio]{Goodfellow_NIPS2014_GAN}
Ian Goodfellow, Jean Pouget-Abadie, Mehdi Mirza, Bing Xu, David Warde-Farley, Sherjil Ozair, Aaron Courville, and Yoshua Bengio.
\newblock Generative adversarial nets.
\newblock In \emph{Advances in Neural Information Processing Systems}. Curran Associates, Inc., 2014.

\bibitem[Gowda et~al.(2022)Gowda, Rohrbach, Keller, and Sevilla-Lara]{Gowda_ECCV2022_Learn2Augment}
Shreyank~N. Gowda, Marcus Rohrbach, Frank Keller, and Laura Sevilla-Lara.
\newblock Learn2augment: Learning to^^c2^^a0composite videos for^^c2^^a0data augmentation in^^c2^^a0action recognition.
\newblock In \emph{Computer Vision -- ECCV 2022}, pages 242--259, Cham, 2022. Springer Nature Switzerland.

\bibitem[Goyal et~al.(2017)Goyal, Ebrahimi~Kahou, Michalski, Materzynska, Westphal, Kim, Haenel, Fruend, Yianilos, Mueller-Freitag, Hoppe, Thurau, Bax, and Memisevic]{Goyal_2017ICCV_ssv2}
Raghav Goyal, Samira Ebrahimi~Kahou, Vincent Michalski, Joanna Materzynska, Susanne Westphal, Heuna Kim, Valentin Haenel, Ingo Fruend, Peter Yianilos, Moritz Mueller-Freitag, Florian Hoppe, Christian Thurau, Ingo Bax, and Roland Memisevic.
\newblock The "something something" video database for learning and evaluating visual common sense.
\newblock In \emph{Proceedings of the IEEE International Conference on Computer Vision (ICCV)}, 2017.

\bibitem[Hashiguchi and Tamaki(2022)]{Hashiguchi_2022_ACCVW_MSCA}
Ryota Hashiguchi and Toru Tamaki.
\newblock Temporal cross-attention for action recognition.
\newblock In \emph{Proceedings of the Asian Conference on Computer Vision (ACCV) Workshops}, pages 276--288, 2022.

\bibitem[He et~al.(2016)He, Shirakabe, Satoh, and Kataoka]{He_ECCVW2016_Action_Recognition_Without_Human}
Yun He, Soma Shirakabe, Yutaka Satoh, and Hirokatsu Kataoka.
\newblock Human action recognition without human.
\newblock In \emph{Computer Vision -- ECCV 2016 Workshops}, pages 11--17, Cham, 2016. Springer International Publishing.

\bibitem[Ho et~al.(2022)Ho, Salimans, Gritsenko, Chan, Norouzi, and Fleet]{Ho_NEURIPS2022_Video_Diffusion_Models}
Jonathan Ho, Tim Salimans, Alexey Gritsenko, William Chan, Mohammad Norouzi, and David~J Fleet.
\newblock Video diffusion models.
\newblock In \emph{Advances in Neural Information Processing Systems}, pages 8633--8646. Curran Associates, Inc., 2022.

\bibitem[Hutchinson and Gadepally(2021)]{Hutchinson_IEEEAccess2021_Action_Recognition_Survey}
Matthew~S. Hutchinson and Vijay~N. Gadepally.
\newblock Video action understanding.
\newblock \emph{IEEE Access}, 9:\penalty0 134611--134637, 2021.

\bibitem[Jabbar et~al.(2021)Jabbar, Li, and Omar]{Jabbar_ACMSurvey2021_GAN_survey}
Abdul Jabbar, Xi Li, and Bourahla Omar.
\newblock A survey on generative adversarial networks: Variants, applications, and training.
\newblock \emph{ACM Computing Surveys}, 54\penalty0 (8), 2021.

\bibitem[Kay et~al.(2017)Kay, Carreira, Simonyan, Zhang, Hillier, Vijayanarasimhan, Viola, Green, Back, Natsev, Suleyman, and Zisserman]{kay_arXiv2017_kinetics400}
Will Kay, Jo{\~{a}}o Carreira, Karen Simonyan, Brian Zhang, Chloe Hillier, Sudheendra Vijayanarasimhan, Fabio Viola, Tim Green, Trevor Back, Paul Natsev, Mustafa Suleyman, and Andrew Zisserman.
\newblock The kinetics human action video dataset.
\newblock \emph{CoRR}, abs/1705.06950, 2017.

\bibitem[Kimata et~al.(2022)Kimata, Nitta, and Tamaki]{Kimata_MMAsia2022_ObjectMix}
Jun Kimata, Tomoya Nitta, and Toru Tamaki.
\newblock Objectmix: Data augmentation by copy-pasting objects in videos for action recognition.
\newblock In \emph{Proceedings of the 4th ACM International Conference on Multimedia in Asia}, New York, NY, USA, 2022. Association for Computing Machinery.

\bibitem[Kingma and Ba(2015)]{Kingma_ICLR2015_Adam}
Diederik~P. Kingma and Jimmy Ba.
\newblock Adam: {A} method for stochastic optimization.
\newblock In \emph{3rd International Conference on Learning Representations, {ICLR} 2015, San Diego, CA, USA, May 7-9, 2015, Conference Track Proceedings}, 2015.

\bibitem[Kirillov et~al.(2019)Kirillov, He, Girshick, Rother, and Dollar]{Kirillov_2019_CVPR_panoptic_segmentation}
Alexander Kirillov, Kaiming He, Ross Girshick, Carsten Rother, and Piotr Dollar.
\newblock Panoptic segmentation.
\newblock In \emph{Proceedings of the IEEE/CVF Conference on Computer Vision and Pattern Recognition (CVPR)}, 2019.

\bibitem[Kong and Fu(2022)]{Kong_IJCV2022_Action_Recognition_Survey}
Yu Kong and Yun Fu.
\newblock Human action recognition and prediction: {A} survey.
\newblock \emph{Int. J. Comput. Vis.}, 130\penalty0 (5):\penalty0 1366--1401, 2022.

\bibitem[Kuehne et~al.(2011)Kuehne, Jhuang, Garrote, Poggio, and Serre]{Kuehne_ICCV2011_HMDB51}
Hildegard Kuehne, Hueihan Jhuang, Est{\'{\i}}baliz Garrote, Tomaso~A. Poggio, and Thomas Serre.
\newblock {HMDB:} {A} large video database for human motion recognition.
\newblock In \emph{{IEEE} International Conference on Computer Vision, {ICCV} 2011, Barcelona, Spain, November 6-13, 2011}, pages 2556--2563. {IEEE} Computer Society, 2011.

\bibitem[Lin et~al.(2019)Lin, Gan, and Han]{Lin_2019ICCV_TSM}
Ji Lin, Chuang Gan, and Song Han.
\newblock Tsm: Temporal shift module for efficient video understanding.
\newblock In \emph{Proceedings of the IEEE/CVF International Conference on Computer Vision (ICCV)}, 2019.

\bibitem[Lin et~al.(2014)Lin, Maire, Belongie, Hays, Perona, Ramanan, Doll{\'{a}}r, and Zitnick]{Lin_ECCV2014_COCO}
Tsung{-}Yi Lin, Michael Maire, Serge~J. Belongie, James Hays, Pietro Perona, Deva Ramanan, Piotr Doll{\'{a}}r, and C.~Lawrence Zitnick.
\newblock Microsoft {COCO:} common objects in context.
\newblock In \emph{Computer Vision - {ECCV} 2014 - 13th European Conference, Zurich, Switzerland, September 6-12, 2014, Proceedings, Part {V}}, pages 740--755. Springer, 2014.

\bibitem[Luo et~al.(2023)Luo, Chen, Zhang, Huang, Wang, Shen, Zhao, Zhou, and Tan]{Luo_2023_CVPR_VideoFusion}
Zhengxiong Luo, Dayou Chen, Yingya Zhang, Yan Huang, Liang Wang, Yujun Shen, Deli Zhao, Jingren Zhou, and Tieniu Tan.
\newblock Videofusion: Decomposed diffusion models for high-quality video generation.
\newblock In \emph{Proceedings of the IEEE/CVF Conference on Computer Vision and Pattern Recognition (CVPR)}, pages 10209--10218, 2023.

\bibitem[Park et~al.(2019)Park, Liu, Wang, and Zhu]{Park_2019_CVPR_SPADE}
Taesung Park, Ming-Yu Liu, Ting-Chun Wang, and Jun-Yan Zhu.
\newblock Semantic image synthesis with spatially-adaptive normalization.
\newblock In \emph{Proceedings of the IEEE/CVF Conference on Computer Vision and Pattern Recognition (CVPR)}, 2019.

\bibitem[Ramesh et~al.(2021)Ramesh, Pavlov, Goh, Gray, Voss, Radford, Chen, and Sutskever]{Ramesh_ICML2021_DALL-E}
Aditya Ramesh, Mikhail Pavlov, Gabriel Goh, Scott Gray, Chelsea Voss, Alec Radford, Mark Chen, and Ilya Sutskever.
\newblock Zero-shot text-to-image generation.
\newblock In \emph{Proceedings of the 38th International Conference on Machine Learning}, pages 8821--8831. PMLR, 2021.

\bibitem[Rombach et~al.(2022)Rombach, Blattmann, Lorenz, Esser, and Ommer]{Rombach_2022_CVPR_stable_diffusion}
Robin Rombach, Andreas Blattmann, Dominik Lorenz, Patrick Esser, and Bj\"orn Ommer.
\newblock High-resolution image synthesis with latent diffusion models.
\newblock In \emph{Proceedings of the IEEE/CVF Conference on Computer Vision and Pattern Recognition (CVPR)}, pages 10684--10695, 2022.

\bibitem[Shorten and Khoshgoftaar(2019)]{Shorten_BigData2019_augmentation_survey}
Connor Shorten and Taghi~M. Khoshgoftaar.
\newblock A survey on image data augmentation for deep learning.
\newblock \emph{Journal of Big Data}, 6:\penalty0 60, 2019.

\bibitem[Soomro et~al.(2012)Soomro, Zamir, and Shah]{Soomro_arXiv2012_UCF101}
Khurram Soomro, Amir~Roshan Zamir, and Mubarak Shah.
\newblock {UCF101:} {A} dataset of 101 human actions classes from videos in the wild.
\newblock \emph{CoRR}, abs/1212.0402, 2012.

\bibitem[Ulhaq et~al.(2022)Ulhaq, Akhtar, Pogrebna, and Mian]{Ulhaq_arXiv2022_Transformers_Action_Recognition_Survey}
Anwaar Ulhaq, Naveed Akhtar, Ganna Pogrebna, and Ajmal Mian.
\newblock Vision transformers for action recognition: {A} survey.
\newblock \emph{CoRR}, abs/2209.05700, 2022.

\bibitem[Wang et~al.(2022)Wang, Zhao, Tang, Luo, and Zeng]{Wang_arXiv2022_ShiftViT}
Guangting Wang, Yucheng Zhao, Chuanxin Tang, Chong Luo, and Wenjun Zeng.
\newblock When shift operation meets vision transformer: An extremely simple alternative to attention mechanism.
\newblock \emph{CoRR}, abs/2201.10801, 2022.

\bibitem[Wang et~al.(2023)Wang, Huang, Zhao, Tong, He, Wang, Wang, and Qiao]{Wang_2023_CVPR_VideoMAE2}
Limin Wang, Bingkun Huang, Zhiyu Zhao, Zhan Tong, Yinan He, Yi Wang, Yali Wang, and Yu Qiao.
\newblock Videomae v2: Scaling video masked autoencoders with dual masking.
\newblock In \emph{Proceedings of the IEEE/CVF Conference on Computer Vision and Pattern Recognition (CVPR)}, pages 14549--14560, 2023.

\bibitem[Wang et~al.(2018{\natexlab{a}})Wang, Liu, Zhu, Liu, Tao, Kautz, and Catanzaro]{Wang_NeurIPS2018_vid2vid}
Ting-Chun Wang, Ming-Yu Liu, Jun-Yan Zhu, Guilin Liu, Andrew Tao, Jan Kautz, and Bryan Catanzaro.
\newblock Video-to-video synthesis.
\newblock In \emph{Advances in Neural Information Processing Systems}. Curran Associates, Inc., 2018{\natexlab{a}}.

\bibitem[Wang et~al.(2018{\natexlab{b}})Wang, Girshick, Gupta, and He]{Wang_2018CVPR_NonLocal_multiview}
Xiaolong Wang, Ross Girshick, Abhinav Gupta, and Kaiming He.
\newblock Non-local neural networks.
\newblock In \emph{Proceedings of the IEEE Conference on Computer Vision and Pattern Recognition (CVPR)}, 2018{\natexlab{b}}.

\bibitem[Weinzaepfel and Rogez(2021)]{Weinzaepfel_IJCV2021_Mimetics_dataset}
Philippe Weinzaepfel and Gr{\'{e}}gory Rogez.
\newblock Mimetics: Towards understanding human actions out of context.
\newblock \emph{International Journal of Computer Vision}, 129\penalty0 (5):\penalty0 1675--1690, 2021.

\bibitem[Wu et~al.(2019)Wu, Chen, Sharma, Pan, Long, and Blumenstein]{Wu_IJCNN_2019_GAN_aug_Action_Recognition}
Di Wu, Junjun Chen, Nabin Sharma, Shirui Pan, Guodong Long, and Michael Blumenstein.
\newblock Adversarial action data augmentation for similar gesture action recognition.
\newblock In \emph{2019 International Joint Conference on Neural Networks (IJCNN)}, pages 1--8, 2019.

\bibitem[Yan et~al.(2021)Yan, Zhang, Abbeel, and Srinivas]{Yan_arXiv2011_VideoGPT}
Wilson Yan, Yunzhi Zhang, Pieter Abbeel, and Aravind Srinivas.
\newblock Videogpt: Video generation using {VQ-VAE} and transformers.
\newblock \emph{CoRR}, abs/2104.10157, 2021.

\bibitem[Yi et~al.(2019)Yi, Walia, and Babyn]{Yi_MedIA2019_GAN_medical_imaging_survey}
Xin Yi, Ekta Walia, and Paul Babyn.
\newblock Generative adversarial network in medical imaging: A review.
\newblock \emph{Medical Image Analysis}, 58:\penalty0 101552, 2019.

\bibitem[Yun et~al.(2019)Yun, Han, Oh, Chun, Choe, and Yoo]{Yun_2019_ICCV_cutmix}
Sangdoo Yun, Dongyoon Han, Seong~Joon Oh, Sanghyuk Chun, Junsuk Choe, and Youngjoon Yoo.
\newblock Cutmix: Regularization strategy to train strong classifiers with localizable features.
\newblock In \emph{Proceedings of the IEEE/CVF International Conference on Computer Vision (ICCV)}, 2019.

\bibitem[Yun et~al.(2020)Yun, Oh, Heo, Han, and Kim]{Yun_arXiv2020_VideoMix}
Sangdoo Yun, Seong~Joon Oh, Byeongho Heo, Dongyoon Han, and Jinhyung Kim.
\newblock Videomix: Rethinking data augmentation for video classification, 2020.

\bibitem[Zhang et~al.(2021)Zhang, Hao, and Ngo]{Zhang_ACMMM2021_TokenShift}
Hao Zhang, Yanbin Hao, and Chong-Wah Ngo.
\newblock Token shift transformer for video classification.
\newblock In \emph{Proceedings of the 29th ACM International Conference on Multimedia}, page 917^^e2^^80^^93925, New York, NY, USA, 2021. Association for Computing Machinery.

\bibitem[Zhang and Agrawala(2023)]{Zhang_arXiv2023_ControlNet}
Lvmin Zhang and Maneesh Agrawala.
\newblock Adding conditional control to text-to-image diffusion models.
\newblock \emph{CoRR}, abs/2302.05543, 2023.

\bibitem[Zhang et~al.(2020)Zhang, Jia, Chen, Zhang, and Yong]{Zhang_ACMMM2020_GAN_aug_Action_Recognition}
Yumeng Zhang, Gaoguo Jia, Li Chen, Mingrui Zhang, and Junhai Yong.
\newblock Self-paced video data augmentation by generative adversarial networks with insufficient samples.
\newblock In \emph{Proceedings of the 28th ACM International Conference on Multimedia}, page 1652^^e2^^80^^931660, New York, NY, USA, 2020. Association for Computing Machinery.

\bibitem[Zhu et~al.(2017)Zhu, Zhang, Pathak, Darrell, Efros, Wang, and Shechtman]{Zhu_NIPS2017_BicycleGAN}
Jun-Yan Zhu, Richard Zhang, Deepak Pathak, Trevor Darrell, Alexei~A Efros, Oliver Wang, and Eli Shechtman.
\newblock Toward multimodal image-to-image translation, 2017.

\bibitem[Zou et~al.(2022)Zou, Choi, Wang, and Huang]{Zou_CVIU2022_ActorCutMix}
Yuliang Zou, Jinwoo Choi, Qitong Wang, and Jia-Bin Huang.
\newblock Learning representational invariances for data-efficient action recognition.
\newblock \emph{Computer Vision and Image Understanding}, page 103597, 2022.

\end{thebibliography}
    }


\end{document}